# Modelling Observation Correlations for Active Exploration and Robust Object Detection

**Javier Velez**　　　　　　　　　　　　　　　　　　　　VELEZJ AT MIT.EDU
**Garrett Hemann**　　　　　　　　　　　　　　　　GHEMANN AT ALUM.MIT.EDU
**Albert S. Huang**　　　　　　　　　　　　　　　　　　ASHUANG AT MIT.EDU
*MIT Computer Science and Artificial Intelligence Laboratory*
*Cambridge, MA, USA*

**Ingmar Posner**　　　　　　　　　　　　　　INGMAR AT ROBOTS.OX.AC.UK
*Mobile Robotics Group*
*Dept. Of Engineering Science, Oxford University*
*Oxford, UK*

**Nicholas Roy**　　　　　　　　　　　　　　　　NICKROY AT CSAIL.MIT.EDU
*MIT Computer Science and Artificial Intelligence Laboratory*
*Cambridge, MA, USA*

## Abstract

Today, mobile robots are expected to carry out increasingly complex tasks in multifarious, real-world environments. Often, the tasks require a certain semantic understanding of the workspace. Consider, for example, spoken instructions from a human collaborator referring to objects of interest; the robot must be able to accurately detect these objects to correctly understand the instructions. However, existing object detection, while competent, is not perfect. In particular, the performance of detection algorithms is commonly sensitive to the position of the sensor relative to the objects in the scene.

This paper presents an online planning algorithm which learns an explicit model of the spatial dependence of object detection and generates plans which maximize the expected performance of the detection, and by extension the overall plan performance. Crucially, the learned sensor model incorporates spatial correlations *between* measurements, capturing the fact that successive measurements taken at the same or nearby locations are *not* independent. We show how this sensor model can be incorporated into an efficient forward search algorithm in the information space of detected objects, allowing the robot to generate motion plans efficiently. We investigate the performance of our approach by addressing the tasks of door and text detection in indoor environments and demonstrate significant improvement in detection performance during task execution over alternative methods in simulated and real robot experiments.

## 1. Introduction

Years of steady progress in mapping and navigation techniques for mobile robots have made it possible for autonomous agents to construct accurate geometric and topological maps of relatively complex environments and to robustly navigate within them (e.g., Newman, Sibley, Smith, Cummins, Harrison, Mei, Posner, Shade, Schroeter, Murphy, Churchill, Cole, & Reid, 2009). Lately, mobile robots have also begun to perform high-level tasks such as the following of natural language instructions or an interaction with a particular object, requiring a relatively sophisticated interpretation by an agent of its workspace. Some of the recent literature therefore focuses on augmenting metric maps with higher-order semantic information such as the location and identity of objects in the workspace (see Fig. 1).

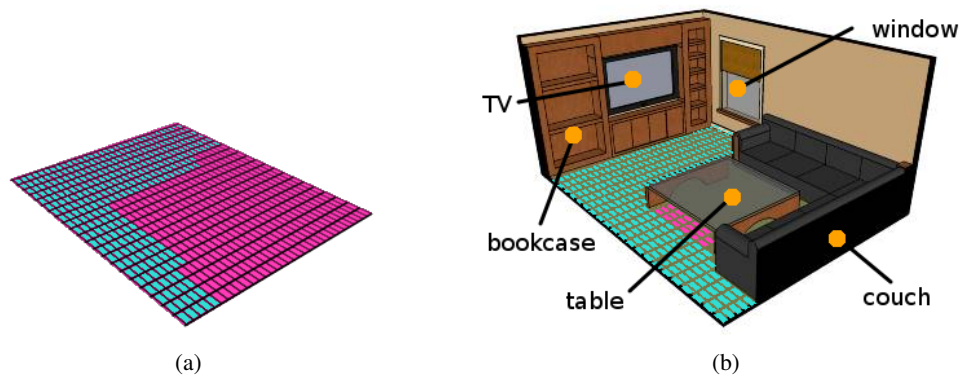

(a)                                              (b)

Figure 1: A traditional geometric environment map (a) represented as a simple two dimensional occupancy grid with regions that are free-space (cyan) and not (red) useful for navigation and localization, (b) the geometric map augmented with semantic information about the identity, structure and location of objects in the world allowing for richer interactions between agent and workspace.

To this end, advances in both vision- and laser-based object detection and recognition have been leveraged to extract semantic information from raw sensor data (e.g., Posner, Cummins, & Newman, 2009; Douillard, Fox, & Ramos, 2008; Martinez-Mozos, Stachniss, & Burgard, 2005; Anguelov, Koller, Parker, & Thrun, 2004). Commonly, the output of such a detection system is accepted *prima facie*, possibly with some threshold on the estimated sensor error. A consequence of directly using the results of an object detector is that the quality of the resulting map strongly depends on the shortcomings of the object detector. Vision-based object detection, for example, is oftentimes plagued by significant performance degradation caused by a variety of factors including a change of aspect compared to that encountered in the training data, changes in illumination and, of course, occlusion (e.g., Coates & Ng, 2010; Mittal & Davis, 2008). Both aspect and occlusions can be addressed naturally by a mobile robot: the robot can choose the location of its sensors carefully before acquiring the data and performing object detection, thereby improving the robustness of the detection process by specifically counteracting known detector issues. Rather than placing the burden of providing perfect detections on the detector itself, the robot can *act* to improve its perception. Rarely, however, is this ability of a mobile robot actually exploited when building a semantic map.

In this paper, we present an online planning algorithm for robot motion that explicitly incorporates a model of the performance of an object detector. We primarily address the problem in the context of a robot exploring an unknown environment with the goal of building a map accurately labeled with the location of semantic objects of interest — here, in particular, we consider doors and textual signs. However, our approach can be applied to any problem where the robot must plan trajectories that depend on the location of objects and landmarks of interest in an environment. We show how our planning approach weighs the benefit of increasing its confidence about a potential semantic entity against the cost of taking a detour to a succession of more suitable vantage point. Fig. 2 gives a cartoon illustration of the problem, where a robot encounters a possible new object

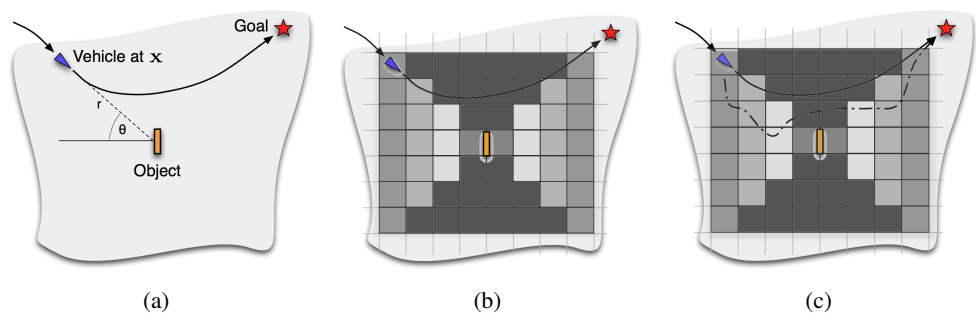

(a)                              (b)                              (c)

Figure 2: A conceptual illustration of (a) the robot at viewpoint $x$ while following the original
trajectory (bold line) towards the goal (red star), (b) the perception field for a particular
object detector centered around an object hypothesis, and (c) an alternative path (bold
dash-dotted line) along a more informative route. Cell shadings indicate the relative value
of observations taken from each cell in terms of mutual information. Lighter values
indicate lower mutual information and therefore desirable vantage points. The challenge
is not only learning how the mutual information varies spatially, but also capturing how
the mutual information at each cell changes with each new measurement.

while executing a path to a goal. Based on the expected information available from possible vantage
points, the robot may decide that the original path provided an accurate model of the object, or it
may choose to modify its path to reduce the possibility of errors in the object model.

We make two primary contributions in this paper. Firstly, we describe a new sensor model
that uses a mixture of Gaussian Processes not only to model the performance of the object detec-
tion system as a function of the robot's relative position to the detected features but to also learn
online a model of how sensor measurements are spatially correlated. Typical estimation and plan-
ning algorithms assume that sensor measurements are conditionally independent of each other given
knowledge of the robot's position, but this assumption is clearly incorrect — properties of the en-
vironment introduce a strong correlation between sensor measurements. Rather than estimate all
possible hidden variables that capture the full sensor model to preserve conditional independence,
we explicitly model spatial correlation of measurements and use this correlation model to esti-
mate the mutual information between measurements taken at different locations. We then use the
mutual information both to bias the random sampling strategy during trajectory generation and to
evaluate the expected cost of each sampled trajectory. Secondly, we show how to incorporate the
learned sensor model into a forward search process using the Posterior Belief Distribution (PBD)
algorithm (He, Brunskill, & Roy, 2010, 2011) to perform computationally efficient deep trajectory
planning. The PBD approximation allows us to compute the expected costs of sensing trajectories
without explicitly integrating over the possible sensor measurements.

While our work is not the first result in actively controlling a sensor to improve its accuracy,
previous work has largely ignored motion cost and has typically assumed observations are condi-
tionally independent given the sensor position. Inspired by recent progress in forward search for
planning under uncertainty, we demonstrate a system which allows us to efficiently find robust ob-
servation plans. This paper builds on our previous work presented at ICAPS 2011 (Velez, Hemann,
Huang, Posner, & Roy, 2011) and provides several substantial extensions. Specifically, we describe
a significantly richer sensor model, extend the approach to an improved planning algorithm and ad-dress an additional object of interest — human-readable text. We demonstrate our overall approach using a real robot as well as simulation studies.

Our exposition begins with the problem formulation of planning trajectories to improve object detection in Section 2. In Section 3 we describe the specific sensor model and how to characterize sensor models using mutual information. Section 4 then gives two different approaches to learning how the sensor models vary spatially, and how observations are correlated spatially. We describe our planning algorithm and how the sensor model is incorporated into the system in Section 5. We follow with a description of the implementation for efficient planning using our sensor models in Section 6. Section 7 describes the object detectors used for our results. Section 8 shows simulation results of how our approach improves object detection compared to other approaches and Section 9 shows the performance of our system in real world trials. In Sections 10 and 11 we conclude with a discussion of related work and future directions.

## 2. Problem Formulation

Consider a robot following a particular trajectory towards a goal in an environment with objects of interest at unknown locations, for example, a rescue robot looking for people in a first-responder scenario. Traditionally, an object detector can be used at waypoints along the trajectory where a detection is either accepted into the map or rejected based on simple detector thresholds. However, the lack of introspection of this approach regarding both the confidence of the object detector and the quality of the data gathered can lead to an unnecessary acceptance of spurious detections. Most systems simply discard lower confidence detections and have no way to improve the estimate with further, targeted measurements. In contrast, we would like the robot to modify its motion to both minimize total travel cost and the cost of errors when deciding whether or not to add newly observed objects to the map.

Let us represent the robot as a point $\mathbf{x} \in \mathbb{R}^2 \times SO(2)$, where $SO(2)$ denotes the special orthogonal group representing orientation and $\mathbb{R}^2$ represents the location in 2D euclidean space. Without loss of generality, we can express a robot trajectory as a set of waypoints $\mathbf{x}^{0:K}$, with an associated motion cost $c_{mot}(\mathbf{x}^{0:K})$ for the sum total travel between waypoints $\mathbf{x}^0$ and $\mathbf{x}^k$. If the robot has a prior map of the environment and is planning a path to some pre-specified goal, then computing a minimum cost path $\mathbf{x}^{0:K}$ is a well-understood motion planning problem.

As the robot moves, it receives output from its object detector that gives rise to a belief over whether a detected object truly exists at the location indicated[1]. We can model the presence of the $i^{th}$ object at some location $(u_i, v_i)$ with the random variable $y_i \in \{\text{object, no-object}\}$. As the system runs, the object detector will fire and give rise to "objects" $Y_i$ at given locations which our system must then reason about and qualify as being either genuine objects or false firings from the object detector.

Let us define a *decision* action $a_i \in \{\text{accept, reject}\}$, where the detected object is either accepted into the map (the detection is determined to correspond to a real object) or rejected (the detection is determined to be spurious). Let us also define a cost $\xi_{dec} : \{\{\text{accept, reject}\} \times \{\text{object, no-object}\}\} \mapsto \mathbb{R}$ for a correct or incorrect accept or reject decision. We cannot know the true cost of the decisions $\{a_i\}$ because we ultimately do not know the true state of objects in the environment. We therefore infer a distribution over the state of each object $p(y)$ and generate a plan to minimize the expected cost $E[\xi_{dec}]$ of individual decision actions given the distribution over objects.

---

1. We assume the robot knows its own location, and has a sufficiently well-calibrated camera to determine the location of the object in the map. In this work, the uncertainty is whether or not an object of a specific type is present at a given location $(u, v)$.

We formulate the planning problem as choosing a plan $\pi$, comprised of a sequence of waypoints and decision actions, $\pi \mapsto \{\mathbf{x}^{0:K} \times a_{0:Q}\}$ for a path of length $k$ and $Q$ hypothesized objects to minimize the total travel cost along the trajectory and the expected costs of the decision actions at the end of the trajectory, such that the optimal plan $\pi^*$ is given by

$$\pi^* = \underset{\mathbf{x}^{0:K},a}{\arg\min} \left( c_{mot}(\mathbf{x}^{0:K}) + c_{det}(\mathbf{x}^{0:K}, a) \right), \tag{1}$$

$$\text{where} \qquad c_{det}(\mathbf{x}^{0:K}, a) = E_{y|\mathbf{x}^{0:K}}[\xi_{dec}(a, y)], \tag{2}$$

where $E_{y|\mathbf{x}^{0:K}}[\cdot]$ denotes the expectation with respect to the robot's knowledge regarding the object, $y$, after having executed path $\mathbf{x}^{0:K}$. The number of hypothesized objects, $Q$, is the number of possible objects the detector fired on after traversing the entire trajectory and is not known beforehand.

Note that the planning problem of computing $\pi^*$ is often formulated as a partially observable Markov decision process or POMDP (Sondik, 1971; Kaelbling, Littman, & Cassandra, 1998), but the POMDP representation will grow with combinatorial complexity in the presence of multiple detections. Furthermore, POMDP solutions assume stationary and Markov model parameters; our sensor model is non-stationary and explicitly non-Markov because we do not want to represent the environmental features that are needed to support a non-Markov sensor model. Since our approach uses a sensor model that adapts with each successive observation, a new POMDP model would need to be constructed and solved after each observation. Lastly, an explicit POMDP model would require the plan to take into account all possible observations the robot might encounter as it carries out the motion trajectory. More precisely, the expected cost of the plan must be computed with respect to all possible observations and objects, rather than just the object distributions. We avoid the resulting computational complexity by using a forward search algorithm similar to forward search approximation techniques for solving POMDPs (Ross, Pineau, Paquet, & Chaib-draa, 2008), which are known to scale well in the presence of complex representations. We also avoid explicitly computing the observation distribution during planning through the use of an approximation technique known as the Posterior Belief Distribution (PBD) algorithm, adapted to our sensor model.

## 3. A Sensor Model for Object Detection

In order to compute the expected cost of decision actions, we must estimate the probability of objects existing in the world given the observations we might see while executing the motion plan. We therefore require a probabilistic model of the object detector that allows us to infer the distribution over the object given measurements, $p(y|z)$. We know that sensor characteristics vary as the robot moves around the object because of interactions with the environment, hence we make this relationship explicit by writing the posterior as $p(y|z, \mathbf{x})$ to include the viewpoint $\mathbf{x}$.

Furthermore, a measurement, $z$, taken from a particular viewpoint $\mathbf{x}$ consists of the output of the object detector, assumed to be a real number indicating the confidence of the detector that an object exists. The distribution over the range of confidence measurements is dependent on a particular object detector and is captured by the random variable $Z$ defined over the continuous range $[z_{min}, z_{max}]$. At every waypoint $\mathbf{x}$ the posterior distribution over $Y$ can be expressed as

$$p(y|z, \mathbf{x}) = \frac{p(z|y, \mathbf{x})p(y)}{\int_{y \in Y} p(z|y, \mathbf{x})p(y)}, \tag{3}$$

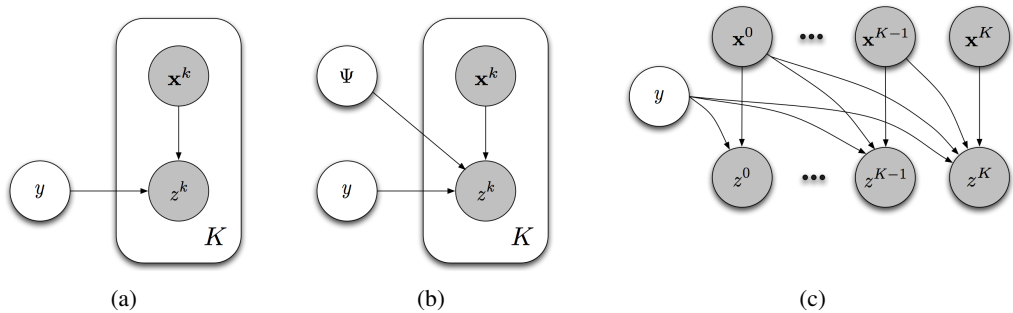

(a)                    (b)                              (c)

Figure 3: Different graphical models representing the observation function. (a) A naive Bayes approximation which assumes every observation $z$ is conditionally independent given knowledge of the object $y$. (b) The true model which assumes observations are independent given knowledge of both the environment $\Psi$ and the object $y$. (c) The model employed here, in which the correlations are approximated by way of a mixture model in the input space of waypoints $\{\mathbf{x} \in \mathbb{R}^2 \times SO(2)\}$ (Equ. 9).

where $p(z|y, \mathbf{x})$ denotes the likelihood, for every possible state of $Y$, of observing a particular detector confidence at $\mathbf{x}$. (The expression would seem to require $p(y|\mathbf{x})$, but $y$ is independent of the waypoint until measurement $z$ is received.)

### 3.1 The Observation Model

Observations $z$ that are directly produced by a physical device, such as a camera, are often treated as conditionally independent given the state of the robot (see Fig. 3a). However, the observations are *not* independent given knowledge only of the current state, but in fact are independent given both the state $Y$ and environment $\Psi$ as shown in Fig. 3(b). If one (or both) of these variables are unknown, then the measurements are no longer first-order Markov but are in fact correlated. This can be seen more intuitively by noting that if the robot were stationary, aimed at a static scene, we would not expect the response of the object detector on successive images to be independent. We anticipate observations from the object detector to be extremely correlated, with the expectation that *no* new information would be gained after more than a handful of images.

To correct our observation model we maintain a history of observations. As more waypoints are visited, knowledge regarding an object can be integrated recursively. Let $\mathcal{T}^K$ denote a trajectory of $K$ waypoint-observation pairs obtained in sequence such that $\mathcal{T}^K = \{(\mathbf{x}^1, z^1), (\mathbf{x}^2, z^2), \ldots, (\mathbf{x}^K, z^K)\}$. Knowledge gained at each step along the trajectory can be integrated into the posterior distribution over $Y$ such that

$$\mathcal{T}^K = (\mathbf{x}^K, z^K) \cup \mathcal{T}^{K-1}, \tag{4}$$

$$p(y|z^K, \mathbf{x}^K, \Psi) \approx p(y|z^K, \mathbf{x}^K, \mathcal{T}^{K-1}), \tag{5}$$

$$= \frac{p(z^K|y, \mathbf{x}^K, \mathcal{T}^{K-1})p(y|\mathcal{T}^{K-1})}{p(z^K|\mathbf{x}^K, \mathcal{T}^{K-1})}, \tag{6}$$

where $z^K$ is the $K^{th}$ observation, which depends not only on the current waypoint but also on the history of measurements and waypoints $\mathcal{T}^{K-1}$. The denominator in Equ. 6 serves to moderate the influence of the measurement likelihood on the posterior based on any correlations existing between observations taken along the trajectory.

The difficulty with the model in Equ. 6 is that the sensor model $p(z^K|y, \mathbf{x}^K, \mathcal{T}^{K-1})$ is difficult to arrive at, depending as it does on the history of measurements. Furthermore, $K$ can be arbitrarily large, so we need a model that predicts observations given an infinite history of observations. We will describe this new sensor model in Section 4.

## 3.2 Perception Fields

Before developing a new sensor model, we first need a way to examine how any sensor model captures the effect of measurements on our posterior belief of the object $y$, and we use the reduction in uncertainty relative to the current belief from the next observation. Given a waypoint $\mathbf{x}^K$ and the trajectory $\mathcal{T}^{K-1}$ visited thus far, the reduction in uncertainty is captured by the mutual information between the object state $y$ and the observation $Z^K$ received at $\mathbf{x}^K$ such that

$$
\begin{aligned}
I(Y, Z^K; \mathbf{x}^K, \mathcal{T}^{K-1}) = \\
H(Y; \mathcal{T}^{K-1}) - H(Y|Z^K; \mathbf{x}^K, \mathcal{T}^{K-1}),
\end{aligned}
\tag{7}
$$

where $H(Y; \mathcal{T}^{K-1})$ and $H(Y|Z^K; \mathbf{x}^K, \mathcal{T}^{K-1})$ denote the entropy and the conditional entropy, respectively (we drop the $\mathbf{x}^K$ from the entropy since the distribution of $Y$ is independent of the robot being at $\mathbf{x}^K$ without the corresponding observation $Z^K$). Thus, $H(Y; \mathcal{T}^{K-1})$ expresses the certainty of the current belief over whether the object exists given the trajectory thus far, unswayed by any new measurements. At every time step, this term is constant for every waypoint considered and is therefore disregarded. The conditional entropy in Equ. 7 can be expanded in terms of the posterior over the state of the hidden variable $Y$ given the previous trajectory $\mathcal{T}^{K-1}$ and an additional measurement taken at $\mathbf{x}^K$, $p(y|z^K, \mathbf{x}^K, \mathcal{T}^{K-1})$ (c.f. Equs. 6 and 9), and the likelihood of $z^K$ taking a particular value conditioned on the trajectory thus far and whether an object viewed from $\mathbf{x}^K$ is present or not, $p(z^K|\mathbf{x}^K, \mathcal{T}^{K-1})$,

$$
\begin{aligned}
H(Y|Z^K; \mathbf{x}^K, \mathcal{T}^{K-1}) = \\
- \int_z \left[ p(z|\mathbf{x}^K, \mathcal{T}^{K-1}) H(Y|z, \mathbf{x}^K, \mathcal{T}^{K-1}) \right],
\end{aligned}
\tag{8}
$$

where $H(Y|z, \mathbf{x}^K, \mathcal{T}^{K-1})$ is computed using the sensor model $p(y|z^K, \mathbf{x}^K, \mathcal{T}^{K-1})$ given in Equ. 6, which is a function of our belief over $y$ after traversing waypoint-observation trajectory $\mathcal{T}^K$. The expected reduction in uncertainty given by the conditional entropy values for all waypoints in the robot's workspace form the *perception field*[2] for a particular object hypothesis (see Fig. 2(b)). We will use the perception field induces by a sensor model in two ways: firstly as a bias in the search for informative path, and secondly as part of the evaluation of the expected cost of each path.

## 4. Correlation Models

As described previously, conventional first-order Markov sensor models do not correctly represent the effect of successive observations that are implicitly correlated by unmodelled environmental variables. The images used by our object detector are not conditionally independent but correlated through the environment $\Psi$. If the robot position and the scene are stationary, then the probability of individual pixel values in successive images will be strongly correlated by the shared environmental representation and robot position, varying only by sensor noise. Subsequently, the object detector responses will also be strongly correlated. However, correctly representing observations in this way requires an environmental model sufficient to capture the image generation process, an intractable computational and modeling burden. Image-based object detectors are not the only detectors which exhibit a dependence on the environment. Any object detector which utilizes properties of the environment (geometric or otherwise) to generate detections cannot *a priori* be treated as producing conditionally independent observations given only the state of the robot. Correctly representing the full generative model of object detection which takes into account all environmental properties used by a detector is frequently an intractable task.

---

2. The reduction in position uncertainty from robot observations across an environment is sometimes known as the "sensor uncertainty field" (Takeda & Latombe, 1992) in active localization. Since our application is object detection, we use the term "perception field" to avoid confusion with the localization problem, but the concepts are otherwise identical.

To overcome this difficulty, we approximate the real process of object detection with a simplistic model of how the images are correlated. We replace the influence of the environment $\Psi$ on correlations between observations with a convex combination of a fully independent model that does not depend on the history of observations, and a correlated observation model that does depend on the history of observations. We treat whether a particular observation is correlated to any previous observation as a random variable. The new posterior belief over the state of the world is computed as

$$p(z^K | y, \mathbf{x}^K, \mathcal{T}^{K-1}) = p(z^K \perp\!\!\!\perp \mathcal{T}^{K-1})p(z^K_{\text{ind}} | y, \mathbf{x}^K)$$
$$+ (1 - p(z^K \perp\!\!\!\perp \mathcal{T}^{K-1}))p(z^K_{\text{corr}} | y, \mathbf{x}^K, \mathcal{T}^{K-1}), \tag{9}$$

where we have marginalized over whether the observation $z^K$ is actually independent from any previous observation or not. We use the notation $A \perp\!\!\!\perp B$ to represent event $A$ is independent of $B$. Factorizing the likelihood in this way (Equ. 9) will allow us to capture the intuition that repeated observations from the similar waypoints add little to the robot's knowledge about the state of the world and should be treated as correlated. Observations from further afield, however, become increasingly independent; $\Psi$ has less of a correlating effect.

In order to have a complete sensor model which uses the factorization in Equ. 9, we need to construct the model for the independent and correlated likelihoods as well as model the probability of a particular detection being independent of any previous detections. The following sections describe two different approaches to modeling the likelihood functions and the probability of independent detections.

## 4.1 The Static Disc Model

Our first sensor model is called the *static disc* sensor model, and is very coarse, assuming that measurements are drawn according to either a learned first-order Markov model or according to the nearest previous observation.

The distribution over the independent detections $z^K_{\text{ind}} | y, \mathbf{x}^K$ is approximated using a histogram-based detector performance on labeled training data. That is, training data is collected by placing the robot at each waypoint in a grid around a training object facing the object. The robot collects a series of images at each waypoint, and generates a histogram of object detection confidences for each waypoint from the collected images, where the histogram gives the probability of a measurement $z^K$ from a specific (relative) waypoint $\mathbf{x}^K$. In contrast, the correlated detection model assumes

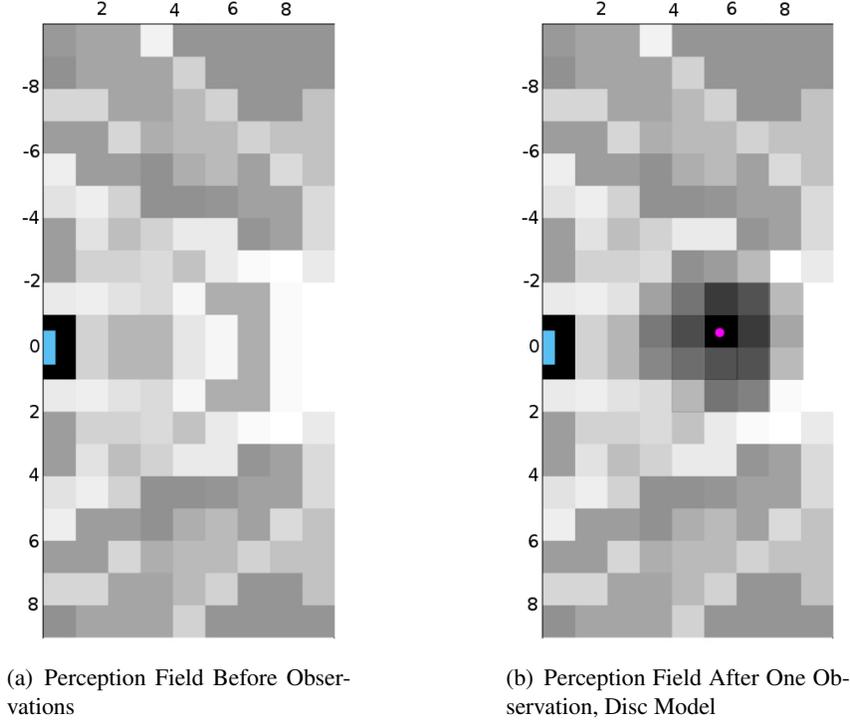

<div align="center">(a) Perception Field Before Observations</div>

<div align="center">(b) Perception Field After One Observation, Disc Model</div>

Figure 4: Perception field for a possible door using the static disc sensor model. The unknown object is at the center (blue) looking towards the right. Brighter regions correspond to waypoints more likely to result in higher confidence posterior beliefs. Observations are taken with the robot at the denoted location (magenta) and oriented to point the sensor directly at the object.

that measurements are fully correlated and always equal to the closest (in $\mathbf{x}$) previously seen observation. As described in Equ. 9 we treat the probability of observation independence as a mixing parameter, $\alpha_{\text{disc}}$ and express it as a truncated linear function of the Euclidean distance, $d$, between two viewpoints. The distribution is normalized with respect to a maximum distance $d_{max}$, beyond which observations are treated as fully independent. Thus,

$$p(z^K \perp\!\!\!\perp \mathcal{T}^{K-1}) = \alpha_{\text{disc}} = \begin{cases} \frac{d}{d_{max}} & \Leftrightarrow d < d_{max} \\ 1 & \Leftrightarrow d \geq d_{max} \end{cases} \quad (10)$$

In other words, no information is gained by taking additional measurements at the same waypoint and the information content of observations increases linearly with distance from previous ones. With reference to Equ. 6, this model results in the belief update,

$$p(y|\mathcal{T}^K) = \left( \alpha_{\text{disc}} \frac{p(z_{\text{ind}}^K|y, \mathbf{x}^K)}{p(z_{\text{ind}}^K|\mathbf{x}^K)} + (1 - \alpha_{\text{disc}}) \right) p(y|\mathcal{T}^{K-1}). \quad (11)$$

Fig. 4 shows two example perception fields for an object detector trained on doors (see Section 7 for the training process). In Fig. 4(a), we see that the highly informative measurements are directly in front of the door, between 8m and 10m. In Fig. 4(b), we see the change to the perception field after an observation. The mixture parameter for our static disc model, $\alpha_{\text{disc}}$, has a $d_{max}$ value empirically chosen to be 3 meters.

## 4.2 The Dynamic Time-Varying Correlation Model

The static disc model shown in the previous section does not allow for the sensor model to change according to the data actively seen during a trajectory. Our purpose in introducing a correlation model is to capture the effect of the environment $\Psi$ on our object detector. An object detector's response to individual object appearances is captured by the dependence on $\Psi$ (for example, a door detector may have a different behavior when detecting highly reflective glass doors versus solid oak doors). However, the static disc model assumes a fixed correlation model and sensor model for all objects of a particular class, regardless of changes in the detector's response across individual instances of object from the same class. The previous model also assumes a strong (truncated) linear relationship between the probability of two observations being correlated and the distance between two observations. We would like to relax this assumption in order to better model a broad range of object detectors. Our second sensor model solves both the aforementioned issues with the static disc model, and also allows for time-varying correlations as more observations are taken of an object. Both sensor models make use of the factorization in Equ. 9, but differ in the models used for the detection likelihoods $p(z_{\text{ind}}^{K}|y, \mathbf{x}^{K})$ and $p(z_{\text{corr}}^{K}|y, \mathbf{x}^{K}, \mathcal{T}^{K-1})$ as well as the structure of $p(z^{K} \perp\!\!\!\perp \mathcal{T}^{K-1})$.

What we would like is a mechanism for learning a correlation between the measurements that can depend on a potentially infinite number of previous measurements, and we use a Gaussian Process (GP) to model both the independent and correlated sensor models. A Gaussian process is a collection of random variables, any finite number of which have a joint Gaussian distribution, and is completely specified by a mean function and a covariance function (Rasmussen & Williams, 2006). We use GP regression in our likelihood models and always use the zero mean function $\mathbf{0}$ and the Squared Exponential (SE) variance function with the following structure:

$$SE(x, x') = \text{sigma} * e^{-(x-x')^T (\text{scale}*I)^{-1}(x-x')/2}. \tag{12}$$

We use the notation $SE_i(X; \theta)$ to mean that kernel $SE_i$ is a function of $X$ and is parameterized by $\theta$.

### 4.2.1 INDEPENDENT AND CORRELATED LIKELIHOOD MODELS

In order to model independent observations we use a Gaussian Process, $\mathcal{GP}_{\text{ind}}$, with zero mean function and squared-exponential covariance function as described above. The kernel parameters, $\theta^{\text{ind}}$, are learned from training data pairs of waypoints $\mathbf{x}$ and observations $z$ as described in section 4.2.3. The GP takes as input a particular waypoint $\mathbf{x}$ and predicts the detector output $z$ at that waypoint. Letting $\mathcal{T}^{\text{train}}$ be the set of labeled waypoint-observation pairs used in $\mathcal{GP}_{\text{ind}}$, the observation model for independent observation becomes

$$z_{\text{ind}}^{K}|y, \mathbf{x}^{K}, \mathcal{T}^{K-1} = z_{\text{ind}}^{K}|y, \mathbf{x}^{K}$$
$$\sim \mathcal{GP}_{\text{ind}}(\mathbf{0}, SE_{\text{ind}}(\mathcal{T}^{\text{train}}, \mathbf{x}^{K}; \theta^{\text{ind}})). \tag{13}$$

where we see that the model depends solely on the training data to provide a prediction.

Similar to the independent model we use a Gaussian Process, $\mathcal{GP}_{\text{corr}}$, with zero mean function and learned SE kernel for the correlated observation model (Equ. 14) trained to model non-independent observations from the object detector. The kernel parameters, $\theta^{\text{corr}}$, are learned from training data as described in Section 4.2.3. Let $\mathcal{T}^{\text{corr-train}}$ be the set of waypoint-observations pairs used to train $\mathcal{GP}_{\text{corr}}$. But, $\mathcal{GP}_{\text{corr}}$ uses the training data *only* to learn the kernel parameters and makes predictions *only* from the data acquired during current trajectory $\mathcal{T}^{K-1}$ so far, which results in the following correlated observation model:

$$z_{\text{corr}}^K | y, \mathbf{x}^K, \mathcal{T}^{K-1} \sim \mathcal{GP}_{\text{corr}}(\mathbf{0}, SE_{\text{corr}}(\mathcal{T}^{K-1}, \mathbf{x}^K; \theta^{\text{corr}})). \tag{14}$$

Unlike our independent model GP which predicts using only training data (Equ. 13), our correlated model GP's predictions are based solely on data observations taken for the current object rather than observation histories from other objects. Predicting the likelihood of $z_{\text{corr}}^K | y, \mathbf{x}^K, \mathcal{T}^{K-1}$ using the GP regression by marginalizing out all but the previous trajectory of observations results in a normal distribution,

$$z_{\text{corr}}^K | y, \mathbf{x}^K, \mathcal{T}^{K-1} \sim N(\mu_{\text{corr},K}, \sigma_{\text{corr},K}^2). \tag{15}$$

The choice to model both independent and correlated observations using GPs results in our overall observation model simplifying to a mixture of two Gaussian distributions,

$$z^K | y, \mathbf{x}^K, \mathcal{T}^{K-1} \sim N(\mu_{obs}, \sigma_{obs}^2). \tag{16}$$

### 4.2.2 MIXTURE PARAMETER AS PROXY FOR INDEPENDENCE

The reason to factor our likelihood into an independent model and a correlated model is to capture the intuition that nearby observations are correlated and are therefore less informative, but we require some baseline model of observations in the remaining robot waypoints. We model the probability of an observation being independent ($p(z^K \perp\!\!\!\perp \mathcal{T}^{K-1})$ in Equ. 9) by treating it as a time-varying spatial mixture parameter $\alpha$. The mixing parameter is chosen to be a function of the variance of the correlation model estimate,

$$p(z^K \perp\!\!\!\perp \mathcal{T}^{K-1}) = p(z^K \perp\!\!\!\perp \mathcal{T}^{K-1} | \mathbf{x}^K, \mathcal{T}^{K-1}) = \alpha(\mathbf{x}^K, \mathcal{T}^{K-1}) = 1 - e^{-\gamma \sigma_{\text{corr},K}^2}. \tag{17}$$

Because we are using an SE kernel function for $\mathcal{GP}_{\text{corr}}$, we know the variance in the prediction $\sigma_{\text{corr},K}$ is a function of the input space distance and is independent of the actual prediction value (Rasmussen & Williams, 2006). Note that $\mathcal{GP}_{\text{corr}}$ is a function of the current trajectory in the world, $\mathcal{T}^{K-1}$, and the current waypoint $\mathbf{x}^K$ but is not a function of the training data $\mathcal{T}^{\text{corr-train}}$. The variance in the estimate from $\mathcal{GP}_{\text{corr}}$ is a function of the distance between the waypoints of observations taken so far for a particular object, and encodes our intuition that observations from similar waypoints are correlated. In fact, as the current waypoint approaches any of the previous observation waypoints, the variance of $z_{\text{corr}}^K | y, \mathbf{x}^K, \mathcal{T}^{K-1}$ approaches 0, $\alpha \to 1$, which means we trust our correlated observation model more than our independent model. Similarly, as the distance between the current waypoint and any previous observation waypoint becomes large, $\alpha \to 0$ and we trust our independent observation model almost exclusively. In other words, little information is gained by taking additional measurements at the same waypoint and the information content of observations increases with distance from previous ones.

As shown in Fig. 3(c), we remove $\Psi$ and add a dependency between previous waypoints and the current observation $z^K$. We use a pair of GPs to model the spatial time-varying properties of correlations in the observation sequence from an object detector.

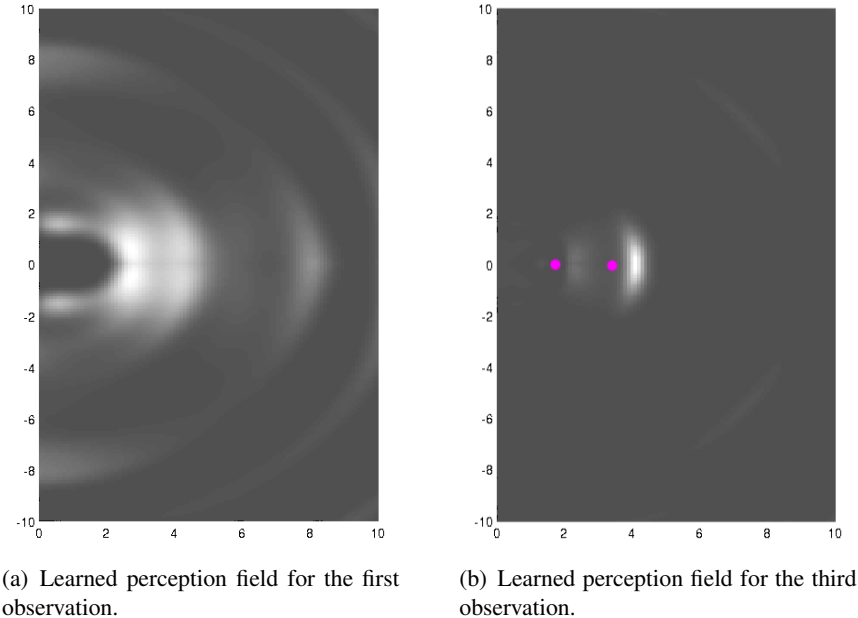

(a) Learned perception field for the first observation.

(b) Learned perception field for the third observation.

Figure 5: Learned perception field for door detector for the (a) first observation and (b) third observation. In (b), the previous observations (shown in magenta) shift where we expect informative vantage points to be. In both panels, the unknown object is centered at the origin facing the right. Brighter regions correspond to waypoints more likely to result in higher confidence posterior beliefs. Observations are taken with the robot at the denoted location (magenta) and oriented to point the sensor directly at the object.

### 4.2.3 TRAINING THE SENSOR MODELS

Our dynamic time-varying observation model consist of a mixture of two Gaussians (see Equ. 16), each of which is modeled using two Gaussian Processes, $\mathcal{GP}_{\text{ind}}$ and $\mathcal{GP}_{\text{corr}}$, for every object hypothesis. Each Gaussian Process maps from locations, $x$, to the resulting object detection score $z$. Every object detector in our system has its own observation model. The independent observation likelihood GPs were trained using all the available training data. Each labeled tuple $(z, \mathbf{x}, y = \{\text{object}, \text{no-object}\})$ was used as if it were an independent sample and fed to the independent GP corresponding to the labeled object state ($\{\text{object}, \text{no-object}\}$). These same training samples were used to learn the SE kernel for the independent GP models. In this way we learn the model of detector output likelihood for the cases when an object truly existed or not, assuming independent observations. These two GPs are shared across all objects and is constant for all measurements.

The correlated observation model GPs have the same learned SE kernel as each other but use different data. The SE kernel is trained only with data from the *same* object since we are trying to learn the model for correlated detections. We split the training data set into subsets which cor-

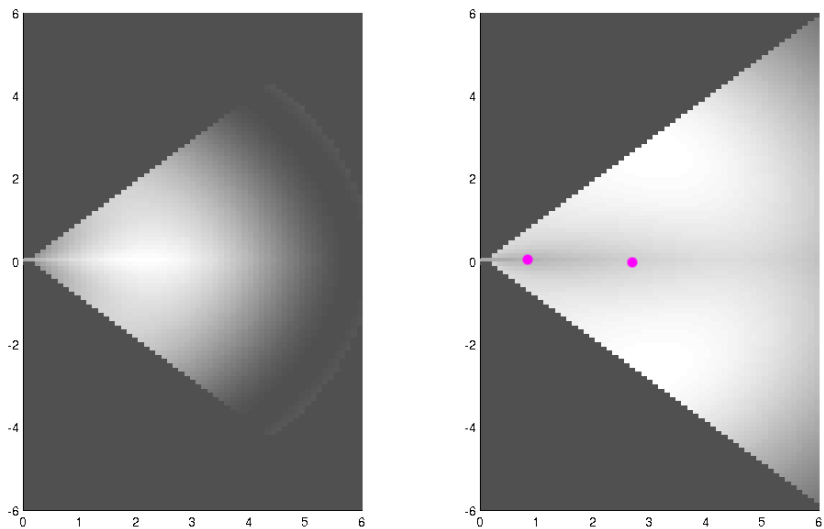

(a) Learned perception field for the first observation.

(b) Learned perception field for the third observation.

Figure 6: Learned perception field for text detector for the (a) first observation and (b) third observation. In (b), the previous observations (shown in magenta) shift where we expect informative vantage points to be. In both panels, the unknown object is centered at the origin facing the right. Brighter regions correspond to waypoints more likely to result in higher confidence posterior beliefs. Observations are taken with the robot at the denoted location (magenta) and oriented to point the sensor directly at the object.

respond to the same objects. The SE kernel parameters are chosen to be the maximal likelihood parameters for the set of subsets. However, once the kernel parameters have been learned, the correlated model GPs are initially devoid of data. These two correlated model GPs are instantiated on a per-object basis and are not shared across objects. Samples are added during runtime while the robot actively observes detector outputs from the world. As such, the correlated model GPs track the current set of waypoints observed for a particular object, whereas the independent model GPs track all of the training samples since they are treated as independent.

Using the learned dynamic time-varying sensor model we derived the initial perception field about a door shown in Fig. 5(a). Fig. 5(b) shows the perception field after several observations have been taken around the door. Notice that the expected amount of information has significantly decreased around the observed points but farther waypoints may still yield useful observations. The initial perception field shows the areas of high expected information gain for an observation according to the training samples for a particular object detector. Since there are not previous observations, the initial perception field shows use the learned independent Gaussian Process for a object detector.

The derived perception field about a text sign is shown in Fig. 6(a). Experimentally, we truncated the text perception field for waypoints that had an aspect of more than 45 degrees to the object for computational efficiency, given that our detector did not fire when viewing signs at more obtuse angles in the training data. Fig. 6(b) shows the perception field after several observations have been taken around the object. Notice that our text detector has a significantly different perception field that our door detector, both in initial shape as well as in response to observations. We see that the door detector has peaks within the perception field, signifying regions with relatively high information gain. The text detector, on the other hand, has a very smooth perception field which drops off mainly as a function of depth.

## 5. Planning To Perceive

Given the sensor model described in the previous section, we now describe a planning algorithm that trades off the necessity of gaining additional information about an object hypothesis against the operational cost of obtaining this information. In particular, when an object is first detected, a new path to the original goal is planned based on the total cost function which includes both the motion cost $c_{mot}$ along the path and the value of measurements from waypoints along the path expressed as a reduction in the expected cost of decision actions. Recall that the cost function consists of two terms: the motion cost $c_{mot}(\mathbf{x}^{0:K})$ and the decision cost $c_{det}(\mathbf{x}^{0:K}, a)$, such that the optimal plan $\pi^*$ is given by Equ. 1, which we reproduce here:

$$\pi^* = \underset{\mathbf{x}^{0:K}, a}{\arg\min} \left( c_{mot}(\mathbf{x}^{0:K}) + c_{det}(\mathbf{x}^{0:K}, a) \right),$$

$$\text{where} \qquad c_{det}(\mathbf{x}^{0:K}, a) = E_{y|\mathbf{x}^{0:K}}[\xi_{dec}(a, y)],$$

where $E_{y|\mathbf{x}^{0:K}}[\cdot]$ denotes the expectation with respect to the robot's knowledge regarding the object, after having executed path $\mathbf{x}^{0:K}$.

### 5.1 Motion cost

The path cost, $c_{mot}(\mathbf{x}^{0:K})$, encompasses operational considerations such as power expended and time taken when moving along a particular trajectory and is typically proportional to the length of the trajectory.

### 5.2 Decision Cost

The decision cost, $c_{det}(\mathbf{x}^{0:K}, a)$, not only captures the expected cost of accepting (or rejecting) a potential object detection, but it also captures the expected yield in information from observations along path $\mathbf{x}^{0:K}$. The trajectory affects the cost of the decision actions in terms of changing the expectation, rather than the decision actions themselves, in effect allowing the algorithm to decide if more observations are needed.

Note that the decision actions can be treated independently of each other and also independently of the robot motion, which allows us to compute the expected decision costs very efficiently. We take advantage of this efficiency to move the minimization over decision actions directly inside the cost function. Abusing notation for $c_{dec}$, we have

$$c_{det}(\mathbf{x}^{0:K}) = \underset{a}{\arg\min}\, c_{det}(\mathbf{x}^{0:K}, a) \qquad (18)$$

$$= \underset{a}{\arg\min}\, E_{y|\mathbf{x}^{0:K}}[\xi_{dec}(a, y)]. \qquad (19)$$

Next, we can write the plan in terms of $\mathbf{x}^{0:K}$.

$$\pi^* = \underset{\mathbf{x}^{0:K}}{\arg\min} \left( c_{mot}(\mathbf{x}^{0:K}) + c_{det}(\mathbf{x}^{0:K}) \right). \tag{20}$$

$\xi_{dec}(\text{accept}, \cdot)$ and $\xi_{dec}(\text{reject}, \cdot)$ are the costs associated with declaring that the object exists or not, respectively, after measuring $z$ at $\mathbf{x}^K$ following traversal of waypoint-observation trajectory $\mathcal{T}^{K-1}$. These costs include the penalties imposed when accepting a true positive detection and when accepting a false positive detection, respectively, and are chosen by the user of the system to reflect the value/penalty of decision for a particular domain.

The expectation inside Equ. 19 relies on a model of $Y$ conditioned on the trajectory $\mathbf{x}^{0:K}$; as can be seen in Fig. 3(c), $Y$ and $\mathbf{x}^{0:K}$ are correlated through $z^K$. During planning, the actual $z^K$ that will be received cannot be known ahead of time, so to evaluate the expectation exactly, it must be taken with respect to both the object state $Y$ and the received observations, as in

$$E_{y|\mathbf{x}^{0:K}}[\xi_{dec}(a, y)] = \tag{21}$$
$$\int_z \left( p(z|\mathbf{x}^K, \mathcal{T}^{K-1}) E_{y|z, \mathbf{x}^{0:K-1}}[\xi_{dec}(a, y)] \right),$$

where $p(z|\mathbf{x}^K, \mathcal{T}^{K-1})$ denotes the probability of obtaining a particular detector confidence value when observing the object from $\mathbf{x}$ given a previous trajectory $\mathcal{T}^{K-1}$, and is computed akin to the posterior in Equ. 6. In Section 6.2 we show how we can efficiently approximate this expectation over the observation sequence by treating our belief as normally distributed.

The planning process proceeds by searching over sequences of $\mathbf{x}^{0:K}$, evaluating paths by approximating the expectations with respect to both the observation sequences and the object state. The paths with the lowest decision cost will tend to be those leading to the lowest posterior entropy, avoiding the large penalty for false positives or negatives.

## 5.3 Multiple Objects

We formally define a *vantage point* relative to an object $y$, $\mathbf{v}_y \in \mathbb{R}^M$, as a vector in an $M$-dimensional feature space describing the configuration of the robot relative to the potential object. We also define a mapping $F : \mathbb{R}^2 \times SO(2) \times Y \mapsto \mathbb{R}^M$ between a robot waypoint $\mathbf{x}$ and its corresponding vantage point $\mathbf{v}_y = F(\mathbf{x}, y)$. In principle, a vantage point need not be restricted to spatial coordinates but may incorporate additional information such as, for example, the degree of occlusion experienced or image contrast (for an appearance based detector). In this work, however, only the range, $r$, and aspect, $\theta$, relative to the object with the robot oriented to directly face the object are considered such that $\mathbf{v}_y \in \mathbb{R} \times SO(2)$ (see Fig. 2a). It is important to note that the system must be able to accurately compute a vantage point; for this paper a stereo camera is used to estimate the distance and orientation of a potential object. The planning approach described so far can be extended to planning in an environment with $Q$ object hypotheses by considering a modified cost function which simply adds the cost for each object. We also augment our $\xi_{dec}(a, y) \rightarrow \xi_{dec}(a, y, i)$ to be able to provide different decision costs for different object types (or even different object instances). The augmentation allows us to specify the relative importance of different objects types in our algorithm. In this work we consider an object's existence to be independent of other objects hence the individual object perception fields are additive for a particular waypoint $\mathbf{x}$. We also restrict ourselves to waypoints which correspond to the robot facing a particular hypothesized object.

Given *no* prior information about object locations, we do not hypothesize how many objects there are in the world. We initially let $Q = 0$ and run the object detector during the robot motion. After each image is processed by the object detector, the system judges whether the detection belongs to an object hypothesis already being considered or not (e.g., using the distance between the hypothesized object and the detection). If the detector determines that the probability of an object at some new location is above a threshold and it does not belong to any hypothesis objects, the number of object hypotheses $Q$ is increased and the robot replans. If the detection is determined to correspond to a particular object hypothesis, the system updates the belief and replans.

### 5.4 Multi-Step Planning

A simple approach for planning considers every possible trajectory to the goal and weights each by the cost of taking the trajectory, choosing the minimum cost trajectory as the plan. This simple algorithm scales approximately exponentially in the length of the planning horizon $T$ and thus rapidly becomes intractable as more observations are considered. We adopt a roadmap scheme in which a fixed number of waypoints are sampled every time a new waypoint is to be added to the current trajectory. A graph is built between the sampled poses, with straight-line edges between samples.

The sampling scheme is biased towards waypoints more likely to lead to useful observations using the perception field (see Section 3.2). Due to the correlations between individual observations made over a trajectory of waypoints, the perception field changes as new observations are added. In particular, the correlation model imposed in this work (Equs. 17 and 9 for the dynamic time-varying model or Eq. 11 for the static disc model) forces

$$\lim_{\# \text{ obs. at } \mathbf{x}^K \to \infty} I(Y, Z^K; \mathbf{x}^K, \mathcal{T}^{K-1}) \to 0,$$

when considering measurements from waypoints already visited. In other words, the robot will prefer to observe the putative object from different waypoints over taking repeated measurements at the same place.

Algorithm REPLANONNEWDETECTION (Fig. 7) summarizes the planned-waypoints approach of sampling and evaluating trajectories to balance increased confidence with motion costs. The algorithm uses the Posterior Belief Distribution framework if able to quickly sample trajectories with many observations, then selects the best of those as the current plan according to our cost metric.

Figure 8 details the stages of our algorithm on an example run where a single door is detected while going towards a goal.

## 6. Efficient Perception Field Computation

Our planning algorithm needs to calculate the perception field for deep planning horizons ($T \gg 1$). The variant of our algorithm which uses our static disc sensor model must evaluate the expected change in our belief over $y$ for every potential future waypoint, and must carry that belief thought each level of our search tree over future trajectories $\mathbf{x}^{K+1:K+T}$. However, when using the dynamic time-varying sensor model we can treat our belief over $y$ as normally distributed. Under the normal distribution approximation, in the limit of an infinite number of observations, the mean of the normal distribution will converge to either 0 or 1 (depending on whether the object is present or not), with

---

**Algorithm** REPLANONNEWDETECTION

---

**Input**: an object detection $z$ at vantage point $\mathbf{x}$

1: *// Step 1: Update Our Belief*
2: **if** using static disc sensor model **then**
3:    $d_{min} = \underset{\mathbf{x}_i \in \mathcal{T}^{K-1}}{\arg\min} |\mathbf{x} - \mathbf{x}_i|$
4:    $\alpha_{\text{disc}} = \frac{d_{min}}{d_{MAX}}$                   *Equ 10*
5:    $p(y) \leftarrow \left(\alpha_{\text{disc}} \frac{p(z^K|y, \mathcal{T}^K)}{p(z^K|\mathcal{T}^K)} + (1 - \alpha_{\text{disc}})\right) p(y|\mathcal{T}^{K-1})$    *Equ 11*
6: **else**
7:    $N(\mu_{\text{corr},K}, \sigma_{\text{corr},K}) \leftarrow \text{PREDICT}(\mathcal{GP}_{\text{corr}}, \mathbf{x})$
8:    $\alpha \leftarrow 1 - e^{-\gamma \sigma_{\text{corr},K}^2}$                    *Equ 17*
9:    $\mu_n \leftarrow \mu_{n-1} + \tilde{K}_n(\tilde{z_n} - W(\mu_{n-1}))$    *Equs 25, 27 and 28*
10:   $\Sigma_n \leftarrow (\Sigma_{n-1} + G_n \ddot{b}_n G_n^T)^{-1}$
11: *// Step 2: Sample Trajectories*
12: $\mathbb{T} \leftarrow \{\}$
13: **while** sampling time remains **do**
14:   $traj \leftarrow \{\}$
15:   **if** using static disc sensor model **then**
16:     $y_0 \leftarrow y$
17:     **for** $i = 1 \rightarrow T$ **do**
18:       $P_i \leftarrow \text{COMPUTE-PERCEPTION-FIELD}(y_{i-1})$    *Equ 8*
19:       $\mathbf{x}_i \sim P_i$ *// sample vantage point*
20:       $p(y_i) \leftarrow E_{z'}[p(z'|\mathbf{x}_i, \mathcal{T}^{K-1}, y_{i-1})p(y_{i-1})]$
21:       $traj \leftarrow traj \cup \mathbf{x}_i$
22:   **else**
23:     $\mu_0' \leftarrow \mu_n$
24:     $\Sigma_0' \leftarrow \Sigma_n$
25:     $\mathcal{GP}_{\text{corr}}^0 \leftarrow \mathcal{GP}_{\text{corr}}$
26:     **for** $i = 1 \rightarrow T$ **do**
27:       $P_i \leftarrow \text{COMPUTE-PERCEPTION-FIELD}(\mu_{i-1}', \Sigma_{i-1}', \mathcal{GP}_{\text{corr}}^{i-1})$    *Equs 8 and 32*
28:       $\mathbf{x}_i \sim P_i$ *// sample vantage point*
29:       $z' \leftarrow \text{PREDICT}(\mathcal{GP}_{\text{corr}}^{i-1}, \mathcal{GP}_{\text{ind}}, \mathbf{x}_i)$    *Equs 9, 13, 14 and 17*
30:       $\mathcal{GP}_{\text{corr}}^i \leftarrow \text{UPDATE-SENSOR-MODEL}(\mathcal{GP}_{\text{corr}}^{i-1}, \mathbf{x}_i, z')$
31:       $\mu_i' \leftarrow \mu_{i-1}' + \tilde{K}_n(\tilde{z_n}' - W(\mu_{i-1}'))$    *Equs 25, 27 and 28*
32:       $\Sigma_i' \leftarrow (\Sigma_{i-1}' + G_n \ddot{b}_n G_n^T)^{-1}$
33:       $traj \leftarrow traj \cup \mathbf{x}_i$
34:   $\mathbb{T} \leftarrow \mathbb{T} \cup traj$
35: EXECUTE-TRAJECTORY $\left( \underset{t' \in \mathbb{T}}{\arg\min} \; \text{COST}(t') \right)$

Figure 7: The waypoint planning algorithm samples trajectories using the perception field, then chooses the trajectories which balance increasing the robot's confidence about an object with minimizing trajectory costs (Equ. 1).

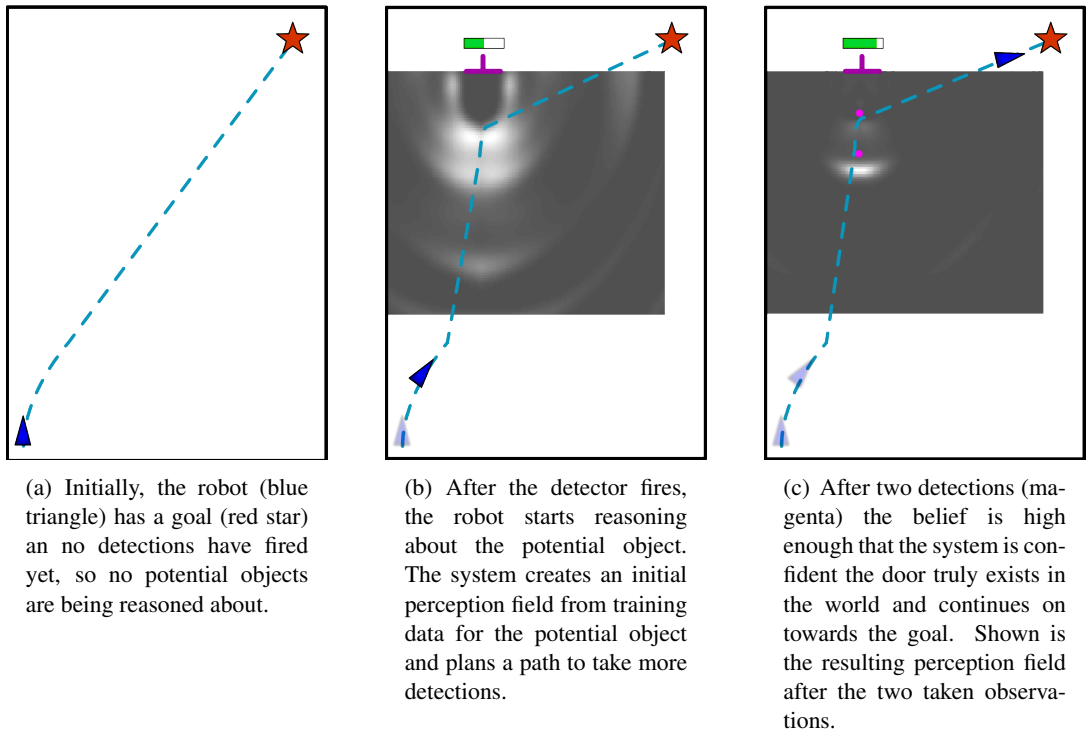

(a) Initially, the robot (blue triangle) has a goal (red star) an no detections have fired yet, so no potential objects are being reasoned about.

(b) After the detector fires, the robot starts reasoning about the potential object. The system creates an initial perception field from training data for the potential object and plans a path to take more detections.

(c) After two detections (magenta) the belief is high enough that the system is confident the door truly exists in the world and continues on towards the goal. Shown is the resulting perception field after the two taken observations.

Figure 8: A sample run of our system, from an initially empty set of objects being reasoned about (a), to a door detector firing and causing a new door object hypothesis and perception field to be created (b). The system then plans and executes a path to the goal which allows it to take advantageous observations of the hypothesized door. After two observations, the system continues towards the goal since the belief of whether the door exists or not is such that the increase in expected reward for further improving the confidence of the object model is not justified by the additional cost (c). Brighter regions of the perception fields correspond to waypoints more likely to result in higher confidence posterior beliefs. Observations are taken with the robot at the denoted location (magenta) and oriented to point the sensor directly at the object. The belief over whether the door truly exists or not is denoted by the green bar.

very small variance. Additionally, the expected cost of a decision will depend only on the variance of the distribution: the smaller the covariance of the normal posterior, the less likely the probability of a decision error. Finally, the posterior covariance of the normal will depend only on the sensor model, and *not* the observation itself. As a result, if we know the sensor model information gain for each measurement, we can predict the posterior covariance, and hence the expected cost of any decision action, without knowing the exact observation sequence itself. This approximation of the binomial measurement function is known as the Posterior Belief Distribution (PBD) algorithm (He et al., 2011), and can be used to efficiently compute the resulting belief after $T$ time steps. We sketch the general idea behind PBD below, which we use to compute the expected entropy reduction in our belief by future observations.

### 6.1 The Belief Over $y$

In reality, an object either exists or does not exists in the world (denoted by $Y$). Labeled training data for the output of an object detector is of the form $(z, \mathbf{x}, y = \{\text{object, no-object}\})$, where we pair the detector output at a particular waypoint with the knowledge of whether the object exists or not. In order to use such labeled samples to train our sensor model (a model of the object detector), we keep track of a belief $\rho$ over whether the object exists or not. Equs. 22 and 23 show our independent and correlated observation model likelihood given $\rho$ using the likelihood given $Y$. We marginalize over our belief for both our independent and correlated observations models (Equs. 13 and 14) to get

$$\begin{aligned}
p(z_{\text{ind}}^K|\rho, \mathbf{x}^K, \mathcal{T}^{K-1}) = {} & \rho \cdot p(z_{\text{ind}}^K|\text{object}, \mathbf{x}^K, \mathcal{T}^{K-1}) \\
& + (1 - \rho) \cdot p(z_{\text{ind}}^K|\text{no-object}, \mathbf{x}^K, \mathcal{T}^{K-1})
\end{aligned} \tag{22}$$

$$\begin{aligned}
p(z_{\text{corr}}^K|\rho, \mathbf{x}^K, \mathcal{T}^{K-1}) = {} & \rho \cdot p(z_{\text{corr}}^K|\text{object}, \mathbf{x}^K, \mathcal{T}^{K-1}) \\
& + (1 - \rho) \cdot p(z_{\text{corr}}^K|\text{no-object}, \mathbf{x}^K, \mathcal{T}^{K-1}),
\end{aligned} \tag{23}$$

where each likelihood is itself modeled as a GP similar to Equ. 13 or 14 for the independent or correlated models respectively.

Noting that we can write the likelihood of $z$ in terms of $\rho$ using $Y$ (Equs. 22 and 23), we can similarly rewrite Equ. 9 in terms of likelihoods based on $\rho$ as

$$\begin{aligned}
p(z^K|\rho, \mathbf{x}^K, \mathcal{T}^{K-1}) = {} & p(z^K \perp\!\!\!\perp \mathcal{T}^{K-1})p(z_{\text{ind}}^K|\rho, \mathbf{x}^K, \mathcal{T}^{K-1}) \\
& + (1 - p(z^K \perp\!\!\!\perp \mathcal{T}^{K-1}))p(z_{\text{corr}}^K|\rho, \mathbf{x}^K, \mathcal{T}^{K-1}).
\end{aligned} \tag{24}$$

### 6.2 Posterior Belief Distribution

The PBD algorithm allows us to estimate the expected information gain from a particular waypoint *without* integrating over potential observations $z$. We begin by framing our problem in the Exponential Family Kalman Filter (efKF) formulation (He et al., 2011) where we treat $\rho$ as the state we are trying to estimate, and we have an exponential family observation model,

$$\rho_n = \rho_{n-1} \sim N(\mu_n, \Sigma_n) \tag{25}$$

$$z_n = exp(z_n \theta_n - b_n(\theta_n) + \kappa_n(z_n)). \tag{26}$$

Given a single observation and the canonical link function $W$ mapping from state to observation parameter $\theta$, the posterior mean and variance of the belief can be computed as,

$$\mu_n = \mu_{n-1} + \tilde{K}_n(\tilde{z_n} - W(\mu_{n-1})) \tag{27}$$

$$\Sigma_n = (\Sigma_{n-1} + G_n \ddot{b_n} G_n^T)^{-1} \tag{28}$$

$$\tilde{K}_n = \Sigma_{n-1} G_n (G_n \Sigma_{n-1} G_n^T + \ddot{b_n}^{-1})^{-1} \tag{29}$$

$$\tilde{z_n} = \theta_n - \ddot{b_n}^{-1} \cdot (\dot{b_n} - z_n) \tag{30}$$

$$G_n = \left. \frac{\partial \theta_n}{\partial \rho_n} \right|_{\rho_n = \mu_{n-1}}. \tag{31}$$

| GP | SE kernel $\sigma$ | SE kernel $l$ |
|---|---|---|
| Text $\mathcal{GP}_{\text{ind}}^{\text{object}}$ | 5.5 | 0.15 |
| Text $\mathcal{GP}_{\text{ind}}^{\text{no-object}}$ | 16.2 | 0.23 |
| Text $\mathcal{GP}_{\text{corr}}^{\text{object}}$ | 4.2 | 0.14 |
| Text $\mathcal{GP}_{\text{corr}}^{\text{no-object}}$ | 4.6 | 0.09 |
| Door $\mathcal{GP}_{\text{ind}}^{\text{object}}$ | 0.52 | 0.58 |
| Door $\mathcal{GP}_{\text{ind}}^{\text{no-object}}$ | 54.3 | 0.49 |
| Door $\mathcal{GP}_{\text{corr}}^{\text{object}}$ | 0.002 | 0.17 |
| Door $\mathcal{GP}_{\text{corr}}^{\text{no-object}}$ | 9303 | 0.31 |

Table 1: The learned GP parameters. Note that the kernel scale for the door and text differ for the correlated observation GPs.

Of particular importance to us is the fact that the posterior covariance has a closed form solution, is independent from the posterior mean (He et al., 2011), and does not require integrating over all possible observations $Z$. We now can compute the posterior covariance after $T$ observations in the future as

$$\Sigma_{n+T} = (\Sigma_{n-1} + \sum_{i=1}^{T} G_i \ddot{b}_i G_i^T)^{-1}. \tag{32}$$

Rather than having to marginalize out potential future observations for every future waypoint, we can compute the variance of our belief after $T$ observations by simply multiplying through the variance of the observations at each of the future waypoints. Given that our perception field is a function of the variance in our belief (since the entropy of a normal distribution is a function of the variance), we can now quickly compute the field for deep observation trajectories. Such efficient computation allows our planning algorithm to sample potential observation trajectories with many observations ($T \gg 1$), thereby increasing the effective search depth of our algorithm and improving our plans.

## 7. Objects: Doors and Signs

Our system is in general agnostic to the type of detector employed and even the sensing modality used. The only constraint is formed by the need to be able to define vantage points (see Section 5.3) to compute a perception field (see Section 3.2). In this work, we chose to test our approach with two different vision-based object detectors: the first leverages the parts-based object detector by Felzenszwalb, Mcallester, and Ramanan (2008) trained to find doors; the second detector aims to spot human-readable text in the world such as commonly found on signs. The use of text-spotting was inspired by the work of Posner, Corke, and Newman (2010) and the authors kindly provided us with a C++ software library of the latest incarnation of their text-spotting engine, which provides detection and parsing facilities for individual words in natural scene images.

The door detector was trained on approximately 1400 positive and 2000 negative examples from manually labeled images collected from a large range of indoor areas excluding our testing environment. Performance on images from the testing environment was low due to false positives triggered by visual structures not present in the training images. The detector could be re-trained to improve performance, but the problem recurs when new environments are encountered. These same examples were also used to train both the sensor models of the door detector.

| *Average* | GREEDY$_{\beta=0.8}$ | GREEDY$_{\beta=0.6}$ | PLANNED$_{\text{disc}}$ | RTBSS |
|---|---|---|---|---|
| Precision | 0.31 ±0.06 | 0.60 ±0.07 | **0.75 ±0.06** | 0.45 ±0.06 |
| Recall | 0.44 ±0.07 | 0.62 ±0.07 | **0.80 ±0.06** | 0.58 ±0.07 |
| Path Length (m) | 67.08 ±2.23 | **41.95 ±0.88** | 54.98 ±3.04 | 47.57 ±0.19 |
| Total Trials | 50 | 50 | 50 | 50 |

Table 2: Simulation performance on single door scenario, with standard error values.

The text detector was trained exactly as described by Posner et al. (2010). The dynamic time-varying sensor model was determined using approximately 1800 positive and 2000 negative examples from manually labeled images collected from an indoor office environment excluding our testing environment. We used the text detector only to localize text in the environment, and did not actually use the contents of the text itself.

For the mixture parameter in our dynamic time-varying sensor model, the scale factor $\gamma$ was chosen as the maximum likelihood estimator using the training data for each detector in the system. We learned scaling values $\gamma_{door} = 6.5$ and $\gamma_{text} = 5.4$. Table 1 shows the learned GP parameters for both the door and text detectors.

## 8. Simulation Results

We first assessed our planning approach using the learned models in a simulated environment. Our simulation environment consisted of a robot navigating through an occupancy map, with object detections triggered according to the learned observation model. We also simulated false positives by placing non-object perceptual features that probabilistically triggered object detections using the learned model for false-alarms. The processing delay incurred by the actual object detector was also simulated (the door detector requires approximately 4.5 seconds to process a spatially decimated 512x384 pixel image while the text detector requires 8 seconds to process a full 1024x768 pixel image).

### 8.1 Comparison Algorithms

For the simulation trials we compared our algorithm against two other algorithms. The GREEDY$_\beta$ algorithm selected the *best* waypoint according to our perception field for each potential object until the belief of each object exceeded a threshold $\beta$. Second, we compared our algorithm against the RTBSS online POMDP algorithm (Paquet, Tobin, & Chaib-draa, 2005). The RTBSS algorithm could not use our full sensor model because of the Markov assumption and only utilized the independent part of the model. One could augment the state space to include the entire history of detections – and therefore use our full sensor model, however such a large state space would render the POMDP intractable in practice. We chose a maximum depth equal to that of our algorithm and modeled the world using a resolution of 2.5 meters for the RTBSS algorithm. We will denote the algorithm using the static disc sensor model as PLANNED$_{\text{disc}}$, and the dynamic time-varying sensor model as PLANNED.

| *Average* | Greedy$_{\beta=0.8}$ | Greedy$_{\beta=0.6}$ | Planned$_{disc}$ | RTBSS |
|---|---|---|---|---|
| Precision | 0.64 ±0.03 | 0.54 ±0.03 | 0.53 ±0.05 | **0.70 ±0.03** |
| Recall | 0.63 ±0.02 | 0.57 ±0.03 | **0.76 ±0.03** | 0.66 ±0.03 |
| Path Length (m) | 153.32 ±4.37 | **121.35 ±1.32** | 138.21 ±7.12 | 160.74 ±6.08 |
| Total Trials | 50 | 50 | 50 | 50 |

Table 3: Simulation performance on multiple door scenario, with standard error values.

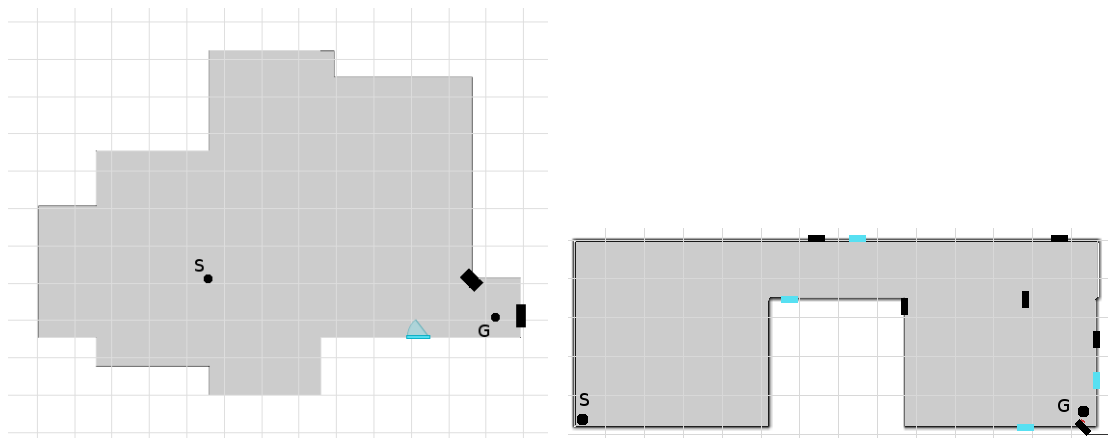

(a) The small simulation environment used for doors containing a single object (blue) and two non-object (black).

(b) The multiple object simulation environment used for doors containing 4 objects (blue) and 6 non-objects (black).

Figure 9: The simulation environments for the static disc sensor model of a door detector.

## 8.2 Static Disc Sensor Model Simulations

First, we tested our Planned$_{disc}$ algorithm on a small the simulation environment with one door object shown in Fig. 9(a). Table 2 shows the simulation results for our static disc model of the door detector. Overall, explicitly planning waypoints resulted in significantly higher performance. The Planned$_{disc}$ algorithm performed better than RTBSS in terms of precision and recall, most likely because our algorithm sampled continuous-space waypoints and the RTBSS algorithm had a fixed discrete representation, while RTBSS paths were shorter.

We evaluated our Planned$_{disc}$ algorithm in a larger, more complex scenario containing four doors and six non-door objects. Fig. 9(b) shows the multiple door simulation environment. Table 3 shows the simulation results for the multi-door scenario. Our Planned$_{disc}$ algorithm resulted in the second shortest paths after Greedy$_{\beta=0.6}$ but with superior detection performance. Planned$_{disc}$ also resulted in significantly shorter paths than RTBSS given the same operating point on the ROC curve.

## 8.3 Dynamic Time-Varying Sensor Model Simulations

We tested our Planned algorithm on both a small simulation with a single text sign and a more complex simulation environment with two signs shown in Figs. 10 and 11. Table 4 shows the results of 20 trials using our text detector sensor model in the single object simulation. For text signs, we see that our deep trajectory planning does not help very much (compare the GREEDY strategy with the Planned strategy which had a planning horizon of 5). The information for the text detector was spread out smoothly (see the perception field Fig. 6(a)) hence the greedy strategy was the best thing to do. However, our planner took into account cost and so resulted in lower precision-recall performance but much shorter path length. We also saw that our correlation sensor model allowed our planned algorithm to perform better than RTBSS. The belief updates predicted by RTBSS were overconfident hence the RTBSS algorithm resulted in shorter path lengths but worse precision-recall performance than our planned-waypoints algorithm.

| *Average* | GREEDY$_{\beta=0.7}$ | PLANNED | RTBSS |
|---|---|---|---|
| Precision | **0.37 ±0.10** | 0.24 ±0.06 | 0.20 ±0.06 |
| Recall | **0.47 ±0.12** | **0.47 ±0.12** | 0.40 ±0.11 |
| Path Length (m) | 35.32 ±1.08 | 20.40 ±0.74 | **18.43 ±0.43** |
| Total Trials | 20 | 20 | 20 |

Table 4: Simulation performance on single sign scenario, with standard error values.

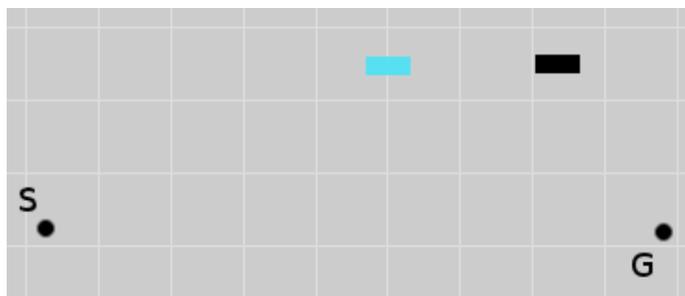

Figure 10: The small simulation environment used for text signs containing a single object (blue) and a single non-object (black).

Next, we evaluated our PLANNED algorithm in a more complex scenario containing two objects and two non-objects shown in Fig. 11. Table 5 shows the simulation results for the multiple-object scenario. The PLANNED algorithm resulted in the best precision-recall performance with short path length. RTBSS also resulted in short path length, but because of the lack of a correlation model became overconfident in its belief, performing significantly worse than the planned-waypoints algorithm in terms of precision-recall.

Fig. 11 also shows the density of all trajectories traversed by each algorithm for all simulations run. Brighter spots denote places where the simulated robot frequented during the simulation runs. We see that the PLANNED algorithm kept the robot close to the shortest path because of our cost function, as does RTBSS. However, our PLANNED algorithm decided to spread detections apart because of the correlation model employed whereas RTBSS over-valued the information gained from nearby observations. The GREEDY algorithm did not take into account the motion cost for taking an observation and so we saw a widespread set of trajectories and waypoints being visited during the simulations.

| *Average* | GREEDY$_{\beta=0.7}$ | PLANNED | RTBSS |
|---|---|---|---|
| Precision | 0.23 ±0.06 | **0.93 ±0.05** | 0.54 ±0.11 |
| Recall | 0.28 ±0.08 | **0.72 ±0.07** | 0.43 ±0.09 |
| Path Length (m) | 66.72 ±1.39 | 33.80 ±1.00 | **23.32 ±0.63** |
| Total Trials | 20 | 20 | 20 |

Table 5: Simulation performance on multiple signs scenario, with standard error values.

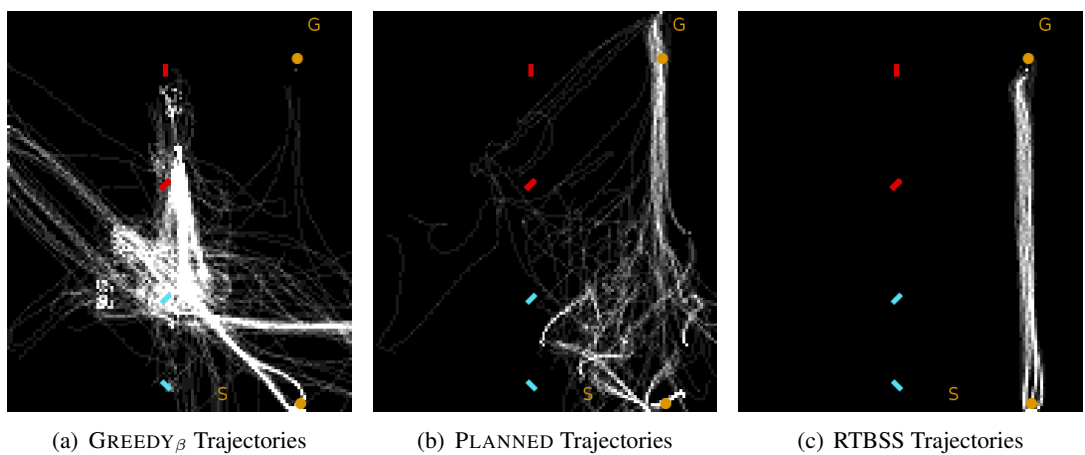

(a) GREEDY$_{\beta}$ Trajectories     (b) PLANNED Trajectories     (c) RTBSS Trajectories

Figure 11: The multiple object simulation environment used for text containing 2 objects (blue) and 2 non-objects (red). Shown are the density of paths taken by the different algorithms during all simulation trials. The planned approach results in a narrower space of paths than GREEDY while avoiding nearby (correlated) observations.

## 8.4 Time Improvements Because of PBD

We ran a comparison between updating our perception field using the PBD algorithm (see Section 6.2) and an update which requires computing the expectation over possible detector outputs. We created a histogram of potential detector values with either 100 or 10 bins and used these sampled to compute the expected mutual information gain (the perception field) for each of those detector output bins. Table 6 shows the results of computing a perception field 100 times. The PBD algorithm allowed us to efficiently calculate the perception field since we did not have to explicitly iterate over the possible detector values but could use Equ. 32.

Updating the perception field was the most time-consuming part of our algorithm since it must be updated when reasoning about future observations during planning. The total run-time was determined by how many trajectories were sampled using the perception field and the depth of these future trajectories, both of which could be tuned to a particular scenario and application. In this paper we let the planning algorithm sample and evaluate trajectories until the same amount of time as running the object detector on a single image had passed.

|     | 100 bins for $z$ | 10 bins for $z$ | PBD |
|-----|------------------|-----------------|-----|
| min | 4.17s            | 1.17s           | 0.85s |
| avg | 4.58s            | 1.20s           | 0.85s |
| max | 4.97s            | 1.22s           | 0.87s |

Table 6: Timing results for computing a perception field using either the PBD algorithm, or explicitly enumerating potential detector values $z$ and computing the expectation over these values.

| *Average* | GREEDY$_{\beta=0.8}$ | PLANNED$_{\text{disc}}$ |
|-----------|----------------------|-------------------------|
| Precision | 0.53 $\pm$0.14 | **0.7 $\pm$0.15** |
| Recall | 0.60 $\pm$0.14 | **0.7 $\pm$0.15** |
| Path Length (m) | 153.86 $\pm$33.34 | **91.68 $\pm$15.56** |
| Total Trials | 10 | 10 |

Table 7: Results of door real-world trials using robot wheelchair, with standard error values.

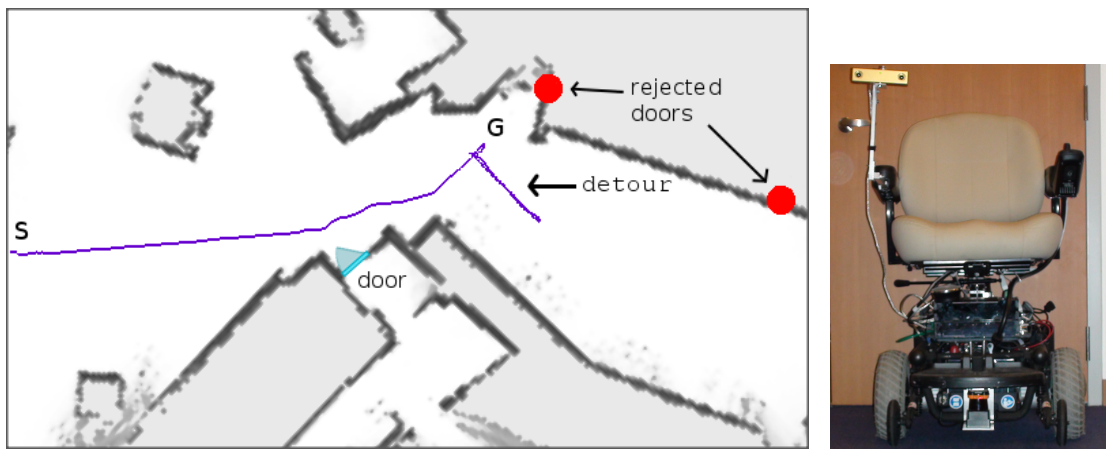

(a) Trajectory executed on the actual robot wheelchair using planned-waypoints from 'S' to 'G' where the robot discovers one true door (cyan). Near the goal, it detects two more possible doors (red dots), detours to inspect them, and (correctly) decides that they are not doors.

(b) Robotic wheelchair platform

Figure 12: Real world trial of door detector using robotic wheelchair platform

## 9. Results For Real World Trials

Finally, we validated the results of the PLANNED$_{\text{disc}}$ and PLANNED algorithms on a robot wheelchair platform (Fig. 12(b)). Our autonomous wheelchair was equipped with onboard laser range scanners, primarily used for obstacle sensing and navigation, a Point Grey Bumblebee2 color stereo camera, and an quad-core laptop as the main processing unit. The stereo camera was used to accurately determine the vantage point for a particular detection. Both door and textual signs were planar, so we fit a plane to the detection bounding box and 3D points from the stereo camera to determine the orientation of the possible object given a detection.

For both door and text real world trials, the robot started at a particular location and orientation. The robot was then given the same goal position such that a nominal trajectory would bring it past one true object (a door or a text sign), and near several fixtures which trigger object detections. Initially, the system had no object hypothesis and the detector was run continuously as it moved towards the goal in the shortest path. As the object detector fired, the system started reasoning about the object hypothesis corresponding to the detections. The robot deviated from the shortest path to take observations for certain object hypothesis as determined by the cost function. Finally, the robot reached the goal and the trial ended. The object hypothesis were accepted if the belief was greater than $0.5$. The cost of an incorrect decision was set to be $16$ times the cost of a meter in path length, and the cost for a correct decision was set to be the negative of an incorrect decision. All trials were capped at $20$ minutes and were done in a real office environment without special accommodations to be as realistic as possible.

Fig. 12(a) shows the location of the door trials. The robot always started at the start location (marked by 'S') and was given the same goal location ('G'). There was a single door which could be seen from the path from start to the goal. Near the goal there were also a set of windows and light fixtures which often caused the door detector to fire. Fig. 12(a) illustrates the trajectory executed during a single trial of the PLANNED$_\text{disc}$ algorithm, and Table 7 summarizes the results of all trials for doors. GREEDY$_{\beta=0.8}$ was chosen as a baseline comparison since it was the best performing of the existing algorithms according to Table 3. Our PLANNED$_\text{disc}$ algorithm resulted in significantly shorter trajectories while maintaining comparable precision and recall. For doors detected with substantial uncertainty, our algorithm planned more advantageous waypoints to increase its confidence and ignored far away detections because of high motion cost. It is interesting to see in Fig. 12(a) how our algorithm deviated to take observations of the false detections near the goal location, ultimately correctly deciding that those object hypothesis were in fact not doors.

We similarly conducted an experiment using the PLANNED algorithm and GREEDY$_{\beta=0.7}$ on the same robotic wheelchair platform with the text detection algorithm. The robot was given a nominal trajectory which brought it past a single textual sign (a poster with an office number on it placed at a common location for poster notifications). The trials were run during the daytime hours to allow for both artificial as well as natural lighting and common environmental changes such as people walking by the robot. Table 8 summarizes the results of 5 real-world trials with both algorithms. We see that the GREEDY algorithm outperformed PLANNED in terms of precision (consistent with our simulation results) but had much longer path lengths. The PLANNED algorithm balanced the cost of gaining new observations against the travel time and resulted in much shorter trajectories. The large path-length associated with the greedy algorithm came from two sources: first, the greedy algorithm did not take path cost into account when deciding the next observation to take, and second the greedy algorithm kept taking pictures of all object hypothesis until the belief was above a certain threshold – this included any sporadic object detections caused by lights or temporary environment noise.

Lastly, we ran a small set of 3 trials using the PLANNED algorithm looking for text signs in a completely different environment than the previous trial. Here we ran the trials at night with very few people walking by. Table 9 shows the results. Even in a different environment, the algorithms behaved similarly, with the GREEDY algorithm outperforming our PLANNED algorithm at the cost of much longer paths.

| *Average* | GREEDY$_{\beta=0.7}$ | PLANNED |
|---|---|---|
| Precision | **1 ±0** | 0.70 ±0.09 |
| Recall | **1 ±0** | 0.80 ±0.09 |
| Path Length (m) | 102.77 ±7.21 | **34.86 ±5.29** |
| Total Trials | 5 | 5 |

Table 8: Results of text real-world trials using robot wheelchair.

| *Average* | GREEDY$_{\beta=0.7}$ | PLANNED |
|---|---|---|
| Precision | **1 ±0** | 0.33 ±0.33 |
| Recall | **1 ±0** | **1 ±0** |
| Path Length (m) | 39.2047 ± 2.76 | **17.5351 ±4.35** |
| Total Trials | 3 | 3 |

Table 9: Results of small text real-world trials in a different location using robot wheelchair.

## 10. Related Work

The problem of planning motion trajectories for a mobile sensor has been explored by a number of domains including planning, sensor placement, active vision and robot exploration. The most general formulation is the partially observable Markov decision process (Sondik, 1971). Exact solutions to POMDPs are computationally intractable, but recent progress has led to approximate solvers that can find good policies for many large, real-world problems (Pineau, Gordon, & Thrun, 2006; Smith & Simmons, 2005; Kurniawati, Hsu, & Lee, 2008; Kurniawati, Du, Hsu, & Lee, 2010). However, the complexity of representing even an approximate POMDP solution has led to forward search strategies for solving POMDPs (Ross et al., 2008; Prentice & Roy, 2009; He et al., 2010). Eidenberger and Scharinger (2010) formulate the problem of choosing sensor locations for active perception as a POMDP very similar in spirit to our formulation. However, they explicitly model the underlying physics of the object generation, model the uncertainty in the object location rather than object type, and are also unable to plan more than one step into the future, and therefore the work is most similar to the GREEDY strategies described in previous sections. Our approach is inspired by the forward search POMDP algorithms, but incorporates a more complex model that approximates the correlations between observations.

In contrast to POMDP models of active sensing, the controls community and the sensor placement community have developed information-theoretic models, where the goal is *only* to minimize a norm of the posterior belief, such as the entropy. This objective function does not depend on the motion costs of the vehicle, and is sub-modular (Krause & Guestrin, 2007). As a consequence, greedy strategies that choose the next-most valuable measurement can be shown to be boundedly close to the optimal, and the challenge is to generate a model that predicts this next-best measurement (Guestrin, Krause, & Singh, 2005; Krause, Leskovec, Guestrin, VanBriesen, & Faloutsos, 2008). In terms of image processing and object recognition, Denzler and Brown (2002) and Sommerlade and Reid (2010) showed that information-theoretic planning could be used to tune camera parameters to improve object recognition performance and applied to multi-camera systems, although their use of exhaustive search over the camera parameters "rapidly becomes unwieldy." Lastly, Sridharan, Wyatt, and Dearden (2008) showed that by formulating an information-theoretic problem as a decision-theoretic POMDP, true multi-step policies did improve the performance of a computer vision system in terms of processing time. However, all of these previous algorithms use models for sequential decision making where the costs of the actions are independent (or negligible), leading to a submodular objective function and limited improvement over greedy strategies.

There has been considerable work in view point selection in active vision which we briefly review here. A few relevant pieces of work include that by Arbel and Ferrie (1999) and more recently Laporte and Arbel (2006) who use a Bayesian approach to model detections that is related to ours, but only searches for the next-best viewpoint, rather than computing a full plan. The work of Deinzer, Denzler, and Niemann (2003) is perhaps most similar to ours in that the viewpoint selection problem is framed using reinforcement learning, but again the authors "neglect costs for camera movement" and identify the absence of costs as a limitation of their work. Similarly, the system by Mittal and Davis (2008) learns a model of object occlusion and uses simulated annealing to solve for the optimal plan; the contribution is to learn a predictive model of good viewpoints. The work by Borotschnig, Paletta, Prantl, and Pinz (2000) uses an appearance-based object detection system to plan viewpoints that minimize the number of observations required to achieve a certain recognition rate, but does not account for correlations in the different observations.

The field of object localization and search has seen some recent advancements. The use of object to object relations seems like a promising direction as shown in the works of Aydemir, Sjöö, Folkesson, Pronobis, and Jensfelt (2011) and Joho, Senk, and Burgard (2011). Our approach differs in that the system uses spatial relations between a single object and multiple observations rather than between different objects. The works by Joho et al. (2011) and Aydemir, Göbelbecker, Pronobis, Sjöö, and Jensfelt (2011) model the environment and achieve good results, whereas our system models the correlation between observations in lieu of modeling the full environment. The idea of "attention" seems a powerful tool for visual search (Tsotsos, 1992) with systems such as those due to Meger, Forssén, Lai, Helmer, McCann, Southey, Baumann, Little, and Lowe (2008) and Andreopoulos, H., Janssen, Hasler, Tsotsos, and Körner (2011) exhibiting excellent results. Rather than using attention, our system utilizes the mutual information and minimizes the cost of taking observations. It is useful to note that while our system minimizes a single cost function which encodes both information and path costs, Ye and Tsotsos (1999) formalized an approach which both maximizes the probability of localizing an object and minimizes the cost.

In robot exploration, where the goal is to generate robot trajectories that learn the most accurate and complete map with minimum travel cost, the costs of motion must be incorporated. Bourgault, Makarenko, Williams, Grocholsky, and Whyte (2002) developed a full exploration planner that incorporated an explicit trade-off between motion plans and map entropy. Stachniss, Grisetti, and Burgard (2005) described a planner that minimized total expected cost, but only performed search over the next-best action. To address the computational challenge, Kollar and Roy (2008) used reinforcement learning to both learn a model over the expected cost to the next viewpoint in the exploration, and minimize the total expected cost of a complete trajectory.

The contribution of our work over the existing work is primarily to describe a planning model that incorporates both action costs and detection errors, and specifically to give an approximate observation model that captures the dynamic correlations between successive measurements that still allows forward-search planning to operate, leading to an efficient multi-step search to improve object detection.

## 11. Conclusion and Future Work

Previous work in planned sensing has largely ignored motion costs of planned trajectories and used simplified sensor models with strong independence assumptions. In this paper, we presented a sensor model that approximates the correlation in observations made from similar vantage points, and an efficient planning algorithm that balances moving to highly informative vantage points with the motion cost of taking detours. We did not fully model the effects of the entire environment on a sensor – an intractable endeavor. Our sensor model simplifies environment interactions by treating them as correlations between the entire history of sensor readings. We placed an emphasis on spatial relations to model the correlations between new sensor readings and the history of previous sensor readings. Because of the properties of Gaussian Processes, our sensor model allows for efficient deep trajectory sampling utilizing the Posterior Belief Distribution framework. We tested our algorithm with two different object detectors (doors and signs) and found better detector dependent observation trajectories than comparable strategies.

The system presented here planned deviations from a particular shortest-path trajectory to a goal in order to detect and localize objects after they had been spotted once. In the future we aim to incorporate a large scale spatial model of where object are likely to be before we have encountered them. Next generation systems will also have to deal with novel objects for which there exists no prior object detector and for whom a detector must be created on the fly. Our goal is to create an end-to-end online adaptive semantic mapping solution which works for arbitrary objects and environments.